\documentclass{article}

\usepackage[preprint]{neurips_2024}

\usepackage[utf8]{inputenc} 
\usepackage[T1]{fontenc}    
\usepackage{hyperref}       
\usepackage{url}            
\usepackage{booktabs}       
\usepackage{amsfonts}       
\usepackage{nicefrac}       
\usepackage{microtype}      
\usepackage{xcolor}         
\usepackage{amsmath}
\usepackage{amssymb}
\usepackage{nomencl}
\usepackage{glossaries}
\usepackage{xspace}
\usepackage{multirow}
\usepackage{graphicx}
\usepackage{bbding}
\usepackage{pifont}
\usepackage{algorithm}
\usepackage{subfigure}
\usepackage{enumitem}
\usepackage{multicol}
\usepackage{caption}
\usepackage{array}
\usepackage{fp}
\usepackage{makecell}
\usepackage{flushend}
\usepackage{cases}
\usepackage{color}
\usepackage{algpseudocode}
\usepackage{listings}
\usepackage{mathrsfs}
\usepackage{longtable}
\usepackage{colortbl}
\usepackage{bm}
\usepackage{caption}

\definecolor{codegreen}{rgb}{0,0.3,0.6}
\definecolor{codegray}{rgb}{0.5,0.5,0.5}

\newcommand{\ie}{\emph{i.e.,}\xspace}
\newcommand{\eg}{\emph{e.g.,}\xspace}

\newcommand{\ignore}[1]{}

\usepackage{tcolorbox}
\definecolor{darkorange}{RGB}{255, 140, 0}
\definecolor{lightgreen}{RGB}{145, 204, 117}
\definecolor{lightyellow}{RGB}{250, 200, 88}
\definecolor{lightred}{RGB}{238, 102, 102}
\definecolor{lightblue}{RGB}{115, 192, 222}
\newtcolorbox{promptbox}[2][Prompt]{
colback=black!5!white,
arc=5pt, 
boxrule=0.5pt,
fonttitle=\bfseries,
title=#1, 
before upper={\scriptsize}, fontupper=\fontfamily{ptm}\selectfont,
colframe=#2, 
}

\title{Imitate, Explore, and Self-Improve: A Reproduction Report on Slow-thinking Reasoning Systems}

\author{%
    Yingqian Min$^{1}$\thanks{Equal contribution.}~,
  Zhipeng Chen$^{1*}$,
  Jinhao Jiang$^{1*}$,
  Jie Chen$^{1}$,\\
  \textbf{
  Jia Deng$^{1}$,
  Yiwen Hu$^{1}$,
  Yiru Tang$^{1}$,
  Jiapeng Wang$^{1}$,
  Xiaoxue Cheng$^{1}$,
  Huatong Song$^{1}$, 
  }\\
  \textbf{Wayne Xin Zhao$^{1}$\thanks{Correspondence to Wayne Xin Zhao.}~, 
  Zheng Liu$^2$,} 
  \textbf{Zhongyuan Wang}$^2$, \textbf{Ji-Rong Wen}$^{1}$
  \\
  $^1$Gaoling School of Artificial Intelligence, Renmin University of China.\\
  $^2$BAAI.\\
  \texttt{\{yingqianm,zhipeng\_chen,jiangjinhao,jrwen\}@ruc.edu.cn}\\
  \texttt{batmanfly@gmail.com}
}

\begin{document}
Technical Report on Slow Thinking with LLMs: II

\maketitle

\begin{abstract}
Recently, slow-thinking reasoning systems, such as o1, have demonstrated remarkable capabilities in solving complex reasoning tasks. These systems typically engage in an extended thinking process before responding to a query, allowing them to generate more thorough, accurate, and well-reasoned solutions.
These systems are primarily developed and maintained by industry, with their core techniques not publicly disclosed. In response, an increasing number of studies from the research community aim to explore the technical foundations underlying these powerful reasoning systems. Building on these prior efforts, this paper presents a reproduction report on implementing o1-like reasoning systems. We introduce an ``imitate, explore, and self-improve'' framework, denoted as \textbf{STILL-2}~\footnote{STILL stands for \underline{S}low \underline{T}h\underline{i}nking with \underline{LL}Ms, and this is the second slow-thinking model developed by our project, ``\emph{Slow Thinking with LLMs}''.}, as our primary technical approach to train the reasoning model.
In the initial phase, we use distilled long-form thought data to fine-tune the reasoning model, enabling it to invoke a slow-thinking mode. The model is then encouraged to explore challenging problems by generating multiple rollouts, which can result in increasingly more high-quality trajectories that lead to correct answers. Furthermore, the model undergoes self-improvement by iteratively refining its training dataset.
To verify the effectiveness of this approach, we conduct extensive experiments on three challenging benchmarks. The experimental results demonstrate that our approach achieves competitive performance compared to industry-level reasoning systems on these benchmarks.
We release our resources at \url{https://github.com/RUCAIBox/Slow_Thinking_with_LLMs}.
\end{abstract}

\section{Introduction}
\label{sec-intro}
Recently, slow-thinking reasoning systems, exemplified by OpenAI's o1~\footnote{https://openai.com/o1/}, have significantly enhanced the capabilities of large language models (LLMs)~\cite{zhao2023survey} in tackling  challenging tasks~\cite{daniel2017thinking,wu2024comparative,zhong2024evaluation,openai2024reasoning}. Unlike previous reasoning approaches~\cite{wei2023chain,Yang2024qwen2.5math,Shao2024DeepSeekMath}, these systems employ test-time scaling, allowing more time for contemplation before responding to a query. This thinking process is also reflected as a text generation process that produces long internal chains of reasoning steps, referred to as \emph{thoughts}, to discover suitable solutions. By examining the generated thought data, we can observe various complex reasoning behaviors exhibited by LLMs, such as planning, divide-and-conquer, self-refinement, summarization, and backtracking. Initially, it may seem surprising that LLMs can manage such complex reasoning processes, even though we know that specific training or inference strategies are employed to support this capability.

To uncover the underlying mechanisms, the research community has been actively exploring slow-thinking reasoning systems and  conducting extensive studies to investigate various potential approaches to reproducing o1-like systems~\cite{jiang2024technical,zhang2024llamaberry,zhang2024o1coder,qin2024o1replicationjourneystrategic,zhao2024marco,k0math,skyworko1}. However, these studies are often limited to specific domains (\eg mathematical domains) or developed using relatively weak base models, which makes the implemented systems significantly inferior to industry systems like o1. 
Implementing an \emph{open} o1-like reasoning system—with all key details publicly disclosed—that can readily generalize across domains and achieve performance comparable to industry-level systems remains a challenging task.

Building on existing research efforts in the literature, our team has been dedicated to advancing the  reproduction of o1-like systems.
To approach this goal, we released a technical report~\cite{jiang2024technical} in November  detailing the implementation of a reasoning framework for addressing mathematical problems. Our framework comprises a policy model, a reward model, and a search algorithm. During inference, the policy model is guided by the reward model to perform the tree search to find correct solutions to mathematical problems. 
We provide an extensive discussion of the explored training and inference methods to implement such a system.

Despite the promising improvements, we quickly realized that the implemented framework in our previous report might not be the correct path toward developing o1-like systems. We identified three major challenges that limit its potential. First, the domain-specific reward model we trained does not generalize well across different domains. Second, performing tree search during the inference stage was very time-consuming, making it impractical for real-world applications. Third, although test-time scaling works, we still cannot achieve train-time scaling to improve model performance. These considerations have led us to reconsider our technical approach to creating o1-like reasoning systems.
 
Our approach is inspired by two main lines of recent progress. Firstly, DeepSeek and Qwen have released the  API or checkpoints for o1-like systems~\cite{deepseek-r1,qwq-32b-preview}, allowing us to closely examine the actual thought processes rather than the summarized versions in o1. This is particularly important for us in obtaining initial labeled data for preliminary attempts. 
Secondly, we have empirically found that fine-tuning LLMs with a small amount of long chain-of-thought data can significantly enhance their performance on complex reasoning tasks, as also reported in previous studies~\cite{qin2024o1replicationjourneystrategic,huang2024o1}. 
Based on these considerations, we speculate that o1 might implement a one-pass decoding process that encompasses both the internal thought and  final solution. In other words, complex reward models and explicit tree search algorithms might not be necessary to support the reasoning process. This speculation has guided our efforts in developing this work for reproducing o1-like systems.

Specifically, we first propose a conceptional framework comprising an  ``\emph{imitate, explore, and self-improve}" process for developing our approach. In the imitation phase, the LLM should learn to tackle tasks by first producing internal thoughts and then generating the solution. Given that this output format significantly differs from a standard response, additional demonstration data is necessary to support this imitation process. This data serves the dual purposes of \emph{format adherence} (\ie following a slow-thinking response) and \emph{ability elicitation} (\ie activating a slow-thinking mode).
In the exploration phase, the LLM should expand its capacity elicited through the demonstration data provided during the imitation stage. We believe it's crucial for the LLM to engage in extensive exploration (typically using techniques like rollout or beam search) on complex tasks to help identify correct solutions to challenging problems. The enhanced outputs generated through exploration are valuable for boosting the model's capabilities. 
Finally, the LLM should leverage the successful trajectories acquired through exploration to further enhance its abilities. It is challenging to continuously obtain training data of higher quality than what the model itself can readily generate, and employing exploration or search methods can help to address this. Once this three-phase training cycle is established, the capabilities of LLMs can be  gradually improved, particularly in handling difficult tasks. 

Following this proposal, in this technical report, we implement an o1-like reasoning system, denoted as \textbf{STILL-2}, which can achieve promising results compared in challenging reasoning tasks.  
Specifically, we collect a small amount of slow-thinking responses from the open o1-like API or checkpoints, and employ these responses as demonstration data to fine-tune our base model. We find that this simple strategy effectively elicits the slow-thinking capacities of LLMs and aligns with the desired output format of both thought and solution. We carefully study how to construct the demonstration dataset by mixing solutions from different domains or with varying levels of difficulty. 
Additionally, we focus on tackling difficult problems for exploration. We employ simple search strategies to obtain correct trajectories (\ie those responses that lead to the ground-truth answers), which are difficult for the fine-tuned model to obtain in a single rollout.
Furthermore, we implement different strategies to achieve self-improvements by either supervised fine-tuning and direct preference optimization. We observe considerable improvements through such a refinement training method. 

To compare our system with industry counterparts, we conduct evaluations on several benchmarks, including MATH-OAI~\cite{LightmanKBEBLLS24lets}, AIME~\footnote{https://huggingface.co/datasets/AI-MO/aimo-validation-amc}, and GPQA~\cite{GPQA}. Experimental results show that when scaling the demonstration instances to 3,900, our variant using distillation-based training even approaches the performance of some industry-level systems. 
Furthermore, our exploration and self-improvement approach also shows very promising results using only 1,100 distilled demonstration instances as seed data. 

\section{Method}
In this section, we provide a detailed introduction of our technical approach to implement o1-like reasoning systems\footnote{Because the exact development of OpenAI's o1 systems is not publicly known, in this paper, ``o1-like'' refers to the reasoning systems that  first conducts extensive reasoning process before producing the final solution. }. We denote the implemented system by \textbf{STILL-2}. 

\begin{figure}[ht]
    \centering    \includegraphics[width=0.9\textwidth]{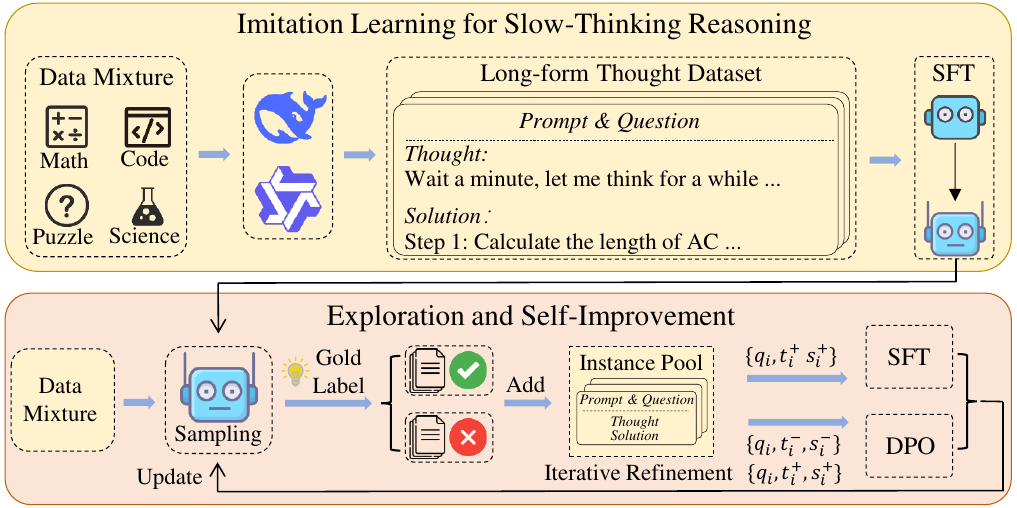}
  \caption{An illustrative overview of our training pipeline for STILL-2.}
  \label{fig:pipline} 
\end{figure}

\subsection{Overview}
 
In this work, we propose a three-phase training approach—\emph{imitate, explore, and self-improve}—to develop reasoning systems similar to o1. After training, the inference phase is also completed by a single-pass text generation process, akin to prior prompt-based methods, with the key distinction that the generated response includes both the reasoning process and the solution. Next, we detail each phase below. 

\begin{itemize}
\item \emph{Imitate}: The core idea is that both the internal thought process and the final solution should be generated in a single response. To achieve this, specific formatting tokens can be used to guide the model in producing such outputs~\cite{Zelikman2024quiet, Herel2024thinking}. We argue that a well-established model, even with a small amount of long-form thought data, can easily adhere to o1-like output formats. This process is fundamentally about following a prescribed format. The key rationale is that, although the entire thought process may be complex, LLMs can effectively handle individual steps (\eg planning, self-refinement, and verification). By using format-following, we can guide LLMs to seamlessly manage and connect these steps. If this hypothesis proves true, two major benefits can be realized: (1) large amounts of data are not needed for format-following, and (2) the approach can be easily generalized to various domains.

\item \emph{Explore}: While imitation enables LLMs to generate o1-like outputs, it may not fully encourage the model to master or improve its ability to use long-form thought to tackle complex tasks. To address this, we believe it is crucial to incorporate exploration, allowing the model to generate progressively better training data on its own. We term this process \emph{exploration}, as the reasoning model cannot directly (or easily) generate a correct solution for challenging tasks. Therefore, search strategies are needed to generate multiple candidate solutions, increasing the likelihood of finding the correct \emph{trajectory}~\cite{wang2023selfconsistency,Yao2023tree} (\ie the entire response consisting of thought and solution). In practice, evaluating the correctness of these attempted trajectories is challenging, requiring a simulated environment with well-trained reward models. In this work, we adopt a simplified method that directly compares the model’s output with the ground-truth answer. Our results show that,  for most of the collected problems, increasing the number of rollouts allows our base model to generate correct trajectories within a reasonable number of attempts.

\item \emph{Self-Improve}: The third phase aims to further enhance the reasoning model’s capabilities by utilizing progressively improved trajectory data. We hypothesize that providing high-quality  demonstrations—particularly those the model cannot easily generate—will effectively strengthen its reasoning abilities. There are several ways to implement this. Typically, we can use rejection sampling for learning with high-quality samples, and direct preference optimization for  learning by comparing high-quality trajectories with lower-quality ones (\eg those that do not lead to the correct answer). Additionally, the ``explore'' and ``self-improve'' phases can be combined through reinforcement learning to achieve systematic model improvement, though this approach generally requires more computational resources and additional training time.

\end{itemize}

We show the overview of our method in Figure~\ref{fig:pipline}. 
Note that this framework is somewhat conceptual, and while we have made some preliminary attempts at instantiating it, our implementation does not fully realize its potential.  
In the following, we will detail the specific implementation of each part in our approach.

\subsection{Imitation Learning for Slow-Thinking Reasoning}\label{sec-imitate}

As discussed in Section~\ref{sec-intro}, we propose using imitation learning to enable the LLM to engage in slow-thinking reasoning—producing an extended process of thought (referred to as \emph{long-form thought}\footnote{We prefer not to use ``chain-of-thought'' since thoughts can be presented flexibly, embodying different reasoning structures.}) before responding to a query. In this section, we will first discuss how to construct the long-form thought dataset for imitation learning~(Section~\ref{sec:data_construction}), and then present the fine-tuning method based on the long-form thought dataset~(Section~\ref{sec:instruction_tuning}).

\subsubsection{Long-form Thought Dataset Construction}
\label{sec:data_construction}
To guide the LLM in producing the long-form thought in a slow-thinking mode followed by the solution, we first need to construct a collection of high-quality demonstration data that exhibits this behavior.

\begin{table}[ht]
    \caption{The summarization of selected data sources.}
    \small
    \centering
    \begin{tabular}{cccc}
        \toprule
        \textbf{Math} & \textbf{Code} & \textbf{Science} & \textbf{Puzzle} \\
        \midrule
        NuminaMath, AIME &Leetcode, OpenCoder&Camel, Gaokao &RiddleSense  \\
        \bottomrule
    \end{tabular}
    \label{tab:data_summarize}
\end{table}

\paragraph{Data Collection.}   In practice, there are three typical approaches to constructing long-form thought data. First, human annotators can be employed to generate this data. Second, LLMs can be employed generate long-form thought data with the assistance of auxiliary search algorithms (\eg Monte Carlo Tree Search). Third, this data can be distilled from o1-like reasoning systems. Given considerations of simplicity and budget constraints, we adopt the third approach for collecting long-form thought data, recognizing that our approach can be extended to include other data collection methods. We access two o1-like reasoning systems: \texttt{DeepSeek-R1-Lite-Preview}~\cite{deepseek-r1} (abbreviated as \emph{R1}) and \texttt{QwQ-32B-preview}~\cite{qwq-32b-preview} (abbreviated as \emph{QwQ}). R1 provides an open API with limited access, while QwQ offers open model checkpoints. We opt not to distill data from o1, as it summarizes internal thought data before presenting it to the user, rendering it not directly suitable for imitation. Upon examining the thought processes of R1 and QwQ, we find that the thoughts can be presented in a relatively flexible and even colloquial manner. To obtain the distilled data, we input the specified problems into both systems and perform multiple rollouts to generate a diverse set of responses. We then automatically label each solution based on whether it leads to the correct answer. 

\begin{figure}[h]
    \centering
    \includegraphics[width=\textwidth]{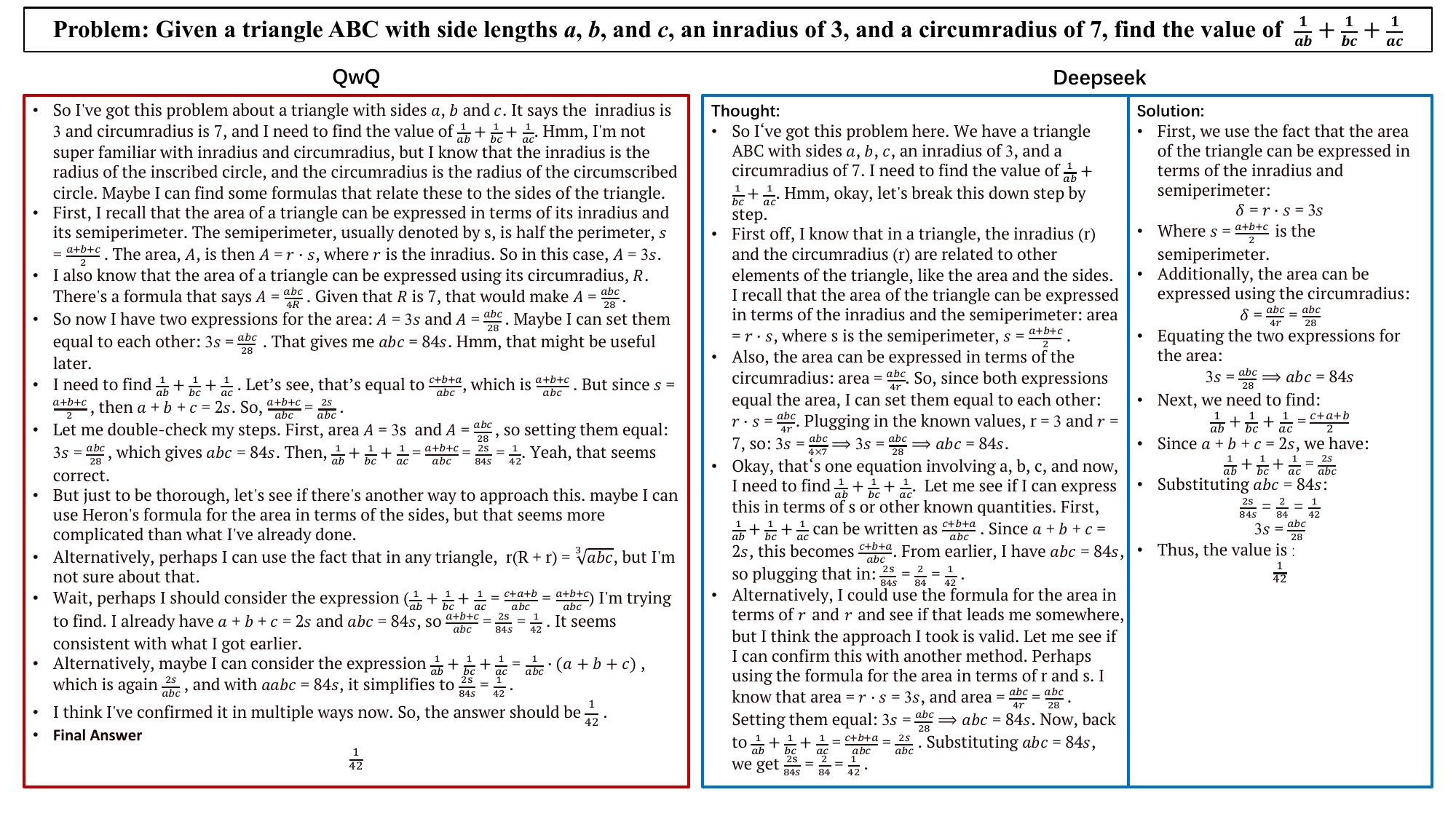}
  \caption{A case study comparing QwQ with DeepSeek in solving math problems.}
  \label{fig:compare} 
\end{figure}
\paragraph{Format Unification.}  After examining the thought processes of R1 and QwQ, we observe that R1 explicitly separates the thinking and solution components, whereas QwQ directly outputs the correct answer following the thinking process. A comparative example of the two systems is shown in Figure~\ref{fig:compare}. 
Following R1's structure, we divide the response into two parts, separated by the tokens ``\texttt{begin\_of\_thought}/\texttt{end\_of\_thought}'' and ``\texttt{begin\_of\_solution}/\texttt{end\_of\_solution}''.
Since QwQ does not explicitly contain a solution component, we consider performing a rollout to complete this part. Specifically, we first fine-tune a reasoning model (either QwQ or our model) with the distilled, formatted data from R1 and then prompt it to complete the solution section. We find that, given the preceding thought process, the reasoning model can readily generate the solution if trained using imitation learning. The final format of our demonstration data is shown below: 

\begin{promptbox}[Long-form Thought Format for Our Reasoning Model]{codegray}
\texttt{<|begin\_of\_thought|>\\
\{different step of thought separated by \textbackslash n\textbackslash n\}\\
<|end\_of\_thought|>\\
\\
<|begin\_of\_solution|>\\
\{formated step-by-step final solution\}\\<|end\_of\_solution|>
}
\end{promptbox}

\paragraph{Data Mixing.} Our goal is to develop more generalized LLMs capable of reasoning across different domains. To achieve this, we begin by mixing demonstration instances (problems paired with their distilled responses) from multiple domains, including mathematics, coding, science, and puzzles. We limit ourselves to these three domains, as we hypothesize that the ability to perform long-form reasoning can transfer easily across them. 
The second consideration is the difficulty of the demonstration instances. Intuitively, applying long-form reasoning to solve relatively simple problems may be less beneficial. Therefore, we focus on collecting more challenging problems from the selected domains. 
Specifically, for the mathematics domain, we select problems from the MATH and Olympiads subsets of the NuminaMATH~\cite{li2024numinamath} dataset, as well as AIME problems collected from the AOPS website~\footnote{https://artofproblemsolving.com} spanning 1983 to 2023. In the coding domain, we use problems labeled as ``Hard'' from the LeetCode website~\footnote{https://leetcode.com}. For the science domain, we gather questions from college entrance examinations and camel-ai~\footnote{https://huggingface.co/camel-ai}, covering subjects such as physics, chemistry, and biology. For the puzzle domain, we select questions from RiddleSense~\footnote{https://github.com/INK-USC/RiddleSense}. We summarize the selected data in Table~\ref{tab:data_summarize}.

\paragraph{Pre-processing Demonstration Data.} After collecting the labeled data, we perform further pre-processing to ensure data quality, including deduplication and filtering. Specifically, when generating long-form thought, existing models often produce issues such as repetitions, gibberish, or mixtures of English and Chinese. To address this, we use rule-based methods (\eg  regex matching and $n$-gram matching) to remove such instances. Another key observation is that longer instances tend to lead to better performance, so we also remove relatively short examples. 
As a result, we obtain a cleaned demonstration dataset suitable for fine-tuning our reasoning model. Additionally, we employ the following prompt to guide the model in performing slow thinking more effectively.

\subsubsection{Long-form Thought Instruction Tuning}
\label{sec:instruction_tuning}
After collecting instruction data for long-form reasoning, we fine-tune the model to replicate the behavior of the slow-thinking mode. Specifically, we first determine the data ratio for each domain through empirical experiments, and then optimize the model using supervised fine-tuning (SFT). The optimization settings are as follows: \texttt{learning rate=1e-5}, and \texttt{batch size=96}.
For the base model, we select \texttt{Qwen2.5-32B-Instruct}, as it has been shown to perform effectively in extensive evaluations. And we utilize the following prompt for instruction tuning.

\begin{promptbox}[Prompt Template for Our Reasoning Model]{codegray}
\texttt{Your role as an assistant involves thoroughly exploring questions through a systematic long thinking process before providing the final precise and accurate solutions. This requires engaging in a comprehensive cycle of analysis, summarizing, exploration, reassessment, reflection, backtracing, and iteration to develop well-considered thinking process.}\\
\\
\texttt{Please structure your response into two main sections: Thought and Solution.}\\
\\
\texttt{In the Thought section, detail your reasoning process using the specified format:}\\
```\\
\texttt{<|begin\_of\_thought|>}\\
\texttt{\{thought with steps separated with "\textbackslash n\textbackslash n"\}}\\
\texttt{<|end\_of\_thought|>\\}
’’’
\\
\texttt{Each step should include detailed considerations such as analisying questions, summarizing relevant findings, brainstorming new ideas, verifying the accuracy of the current steps, refining any errors, and revisiting previous steps.} \\
\\
\texttt{In the Solution section, based on various attempts, explorations, and reflections from the Thought section, systematically present the final solution that you deem correct. The solution should remain a logical, accurate, concise expression style and detail necessary step needed to reach the conclusion, formatted as follows:}\\
```\\
\texttt{<|begin\_of\_solution|>\\
\{final formatted, precise, and clear solution\}\\
<|end\_of\_solution|>}\\
’’’
\\
\texttt{Now, try to solve the following question through the above guidelines:
}
\end{promptbox}

Although we can distill a large amount of instruction data, we retain only several thousand demonstration instances during SFT. Our ultimate goal is to assess the effectiveness of self-improvement learning within this approach. 
In our experiments, we empirically find that capable LLMs can readily learn to perform long-form thinking, and this ability can be transferred across domains. Further discussions and experimental details are provided in Section~\ref{sec-IL}.
\subsection{Exploration and Self-Improvement}
\label{sec-explore}
Although we can increasingly annotate or distill more demonstration data, the process remains largely limited by extra efforts in producing the long-form thought data. In this section, we propose enabling LLMs to explore on their own, gradually generating more data for self-improvement. First, we guide the LLM to explore challenging problems (Section~\ref{sec-explore}), then identify the trajectories that are  suitable for the LLM's learning process (Section~\ref{sec-selection}), and finally use the selected trajectories to further enhance the LLM's reasoning abilities (Section~\ref{sec-self_improve}).

\subsubsection{Exploration on Hard Problems}
\label{sec-explore}

Our main idea is to collect correct trajectories (consisting of both thoughts and solutions) to train our reasoning model. Since we do not include a trainable reward model, we focus on collecting a variety of problems with ground-truth answers for exploration. Specifically, for each problem, we perform multiple rollouts to generate candidate trajectories. This process continues until a solution containing the correct answer is produced. In this way, as we scale the number of rollouts, we can collect an increasing number of problems, while the quality of the trajectories improves as the reasoning model is further trained. This iterative process is crucial for self-improvement training.

We empirically find that challenging problems, which require longer thought processes for reasoning, are particularly useful for improving the model's  performance. In contrast, simpler problems often do not contribute to slow-thinking reasoning and can even degrade model performance. 
Among the hard problems in our collected datasets, the mathematical domain contains a higher proportion, exemplified by the challenging problems from the  Mathematical Olympiad. 
Another interesting observation is that long-form thinking appears to be an inherent capability of LLMs, not limited to specific domains. Even when trained exclusively on mathematical problems, the model can effectively reason in a slow-thinking mode across other domains. It is important to note that the number of hard problems is highly limited, so our training set will be relatively small in scale.

\subsubsection{Iteratively Refined Training Data}
\label{sec-selection}
We propose using iterative training to enhance the slow-thinking capabilities of our model, with the key idea being to generate progressively refined training datasets. This refinement can be approached from two main aspects. First, the dataset can be refined by incorporating more correct trajectories from challenging problems. Second, it can be refined by adding more high-quality trajectories generated by an improved reasoning model.

Specifically, let $\mathcal{D}_0$
  denote the original dataset, consisting of distilled trajectories from external reasoning systems, which is used to train our initial reasoning model. Once the model is trained, we use it to perform exploration and generate additional trajectories. These new trajectories are then added to  $\mathcal{D}_0$, resulting in a new dataset $\mathcal{D}_1$. This process can be repeated iteratively by alternating between training stronger models and generating refined training data. In this way, we can continuously improve the training dataset as the reasoning model evolves.

At each refinement step, we also perform strict pre-processing to filter out low-quality trajectories, such as those that are short or noisy. Additionally, we find that perplexity can serve as a useful metric for data selection~\cite{Ankner2024perplexity}, allowing us to identify and retain more challenging trajectories as recognized by the current reasoning model.

However, as discussed above, a significant limitation is the scarcity of challenging problems, especially those paired with ground-truth answers. As a result, the pool of such problems will be quickly exhausted after only a few iterations. We plan to address this limitation in future work.

\subsubsection{Optimization for Self-improvement}
\label{sec-self_improve}

After discussing how to generate iteratively refined training data, we now introduce the optimization methods for self-improvement. Our aim is to study how exploration can enhance the reasoning capabilities of the models. To achieve this, we apply two straightforward optimization strategies, integrating the refined training datasets: supervised fine-tuning and direct preference optimization. 

\paragraph{Supervised Fine-tuning.} We first consider using SFT. Since we employ length and perplexity as selection metrics to filter out low-quality rollouts, this approach can also be viewed as rejection sampling~\cite{Yuan2023rft,Zelikman2022star}. We adopt the capable model \texttt{Qwen2.5-32B-Instruct}~\cite{qwen2.5} as the base model, denoted as $\mathcal{M}_0$. 
At the $t$-th iteration, $\mathcal{M}_0$ is firstly trained on the previous dataset $\mathcal{D}_{t-1}$, resulting in the improved model $\mathcal{M}_t$. This alternating process of generation and training is repeated multiple times, until our problem pool is exhausted or the maximum number of iterations is reached. 
Note that another training method is to train $\mathcal{M}_t$ based on $\mathcal{M}_{t-1}$. However, this would not benefit the optimization in our experiments, and we speculate that the training set $\mathcal{D}_{t}$ is relatively small in scale, which may even lead to performance degradation in incremental training.

\paragraph{Direct Preference Optimization.} 
Another approach to improving the reasoning model is through direct preference optimization (DPO)~\cite{rafailov2024direct}. For DPO, we need to select paired positive and negative instances for contrastive learning. As mentioned earlier, {we select the correct responses with a higher perplexity score as positive instances and the incorrect responses with a lower perplexity score as negative instances}, which enhances the discrimination difficulty for the reasoning model, allowing it to improve to a greater extent.
Similar to the SFT method, at the $t$-th iteration, we  take the checkpoint $\mathcal{M}_1$ (with the first-round training) as the base model for DPO training.  Additionally, we incorporate an SFT loss to help stabilize the DPO training, using the same problem set. When using DPO, a straightforward method is to align the entire response. An alternative approach is to align only the thought part. As mentioned earlier, we observed that once the thought part is generated, the randomness in generating the solution part becomes quite limited. In other words, a detailed thought process often leads to a relatively certain solution. 
We will examine the effects of aligning different parts in Section~\ref{exp-dpo}.

In addition to the two methods described above, another promising training approach is reinforcement learning~\cite{ppo,ye2024online}, where the policy model is directly trained during the exploration process. However, due to computational resource constraints, we leave this approach for future work.

\section{Experiments}

In this section, we conduct experiments to examine the effectiveness of the implemented framework.  

\subsection{Evaluation Setup}
To demonstrate the effectiveness of our framework, we mainly conduct experiments on three challenging benchmarks: MATH-OAI~\cite{LightmanKBEBLLS24lets}, AIME2024~\footnote{https://huggingface.co/datasets/AI-MO/aimo-validation-amc}, and GPQA~\cite{GPQA}. MATH-OAI contains 500 competition mathematics problems from the MATH~\cite{dan2021math} test set. AIME2024 features 30 problems specifically designed to challenge top high school students with complex problem solving tasks. GPQA consists of 198 multiple-choice problems in biology, physics, and chemistry.  In our experiments, we focus on mathematics as the primary domain, with biology, physics, and chemistry serving as auxiliary domains. Among the math benchmarks, MATH-OAI is considered relatively easier, while AIME2024 is regarded as very challenging. Additionally, due to the small number of test samples in AIME2024, its performance tends to fluctuate in our experiments.

We select \texttt{Qwen2.5-32B-Instruct}~\cite{qwen2.5} as the backbone model because it demonstrates sufficient foundational capabilities to effectively engage in extended reasoning process.
As for baselines, we select several leading o1-like models for comparison~(\ie o1-preview~\cite{openai2024reasoning}, DeepSeek-R1-Lite-Preview~\cite{deepseek-r1}, and QwQ-32B~\cite{qwq-32b-preview}).
In addition, we include GPT-4o~\cite{openai2024gpt4o} and Claude 3.5 Sonnet~\cite{claude3.5}, which are advanced general-purpose models.
We use greedy search to evaluate the performance of our model with maximum tokens set to 32k.

\begin{table}[htbp]
    \centering
    \small
    \setlength\tabcolsep{2.4pt}
    \label{tab:main_res}
    \caption{Performance comparison of different methods on three representative benchmarks. ``Backbone'' refers to CoT reasoning method based on the \texttt{Qwen2.5-32B-Instruct} model, while ``\emph{w/} SFT'' and ``\emph{w/} SFT \& DPO'' denote training with our proposed method. 
    The columns of ``Distill'' and ``Explore'' indicate that the source of training instances, either distillation from R1 and QwQ or exploration by the model itself. 
     The \textbf{bold} fonts denote the best performance among our training variants, and we report the gain over the backbone model  (in percentage). }
      \begin{tabular}{l | cc |cccccccc}
      \toprule
      \textbf{Method}& \multicolumn{2}{c}{\textbf{Num. Data}} & \multicolumn{2}{c}{\textbf{MATH-OAI}} & \multicolumn{2}{c}{\textbf{AIME}} & \multicolumn{2}{c}{\textbf{GPQA}} \\ 
      \cmidrule(r){2-3}\cmidrule(r){4-5}\cmidrule(r){6-7}\cmidrule(r){8-9}
       & \textbf{Distill} & \textbf{Explore} & \textbf{Acc (\%)} & \textbf{Gain (\%)} & \textbf{Acc (\%)} & \textbf{Gain (\%)} & \textbf{Acc (\%)} & \textbf{Gain (\%)} \\
       \midrule
    GPT-4o & - & - & 76.6 & - & 9.3 & - & 53.6 & - \\
    Claude 3.5 Sonnet & - & - & 78.3 & - & 16.0 & - & 65.0 & - \\
      o1-preview & - & - & 85.5 & - & 44.6 & - & 72.3 & - \\
      DeepSeek-R1-Lite-P & - & - & 91.6  & - &  52.5 & - & 58.5 & - \\
      QwQ-32B-preview & - & - & 90.6 & - &  50.0 & - & 65.2 & -\\
      \midrule
      \midrule
       Backbone & - & - & 80.0 & - &  13.3 & - & 43.4 & - \\
       \emph{w/} SFT (STILL-2) & 3.9k & - & \textbf{90.2} & \textbf{+12.8} & \textbf{46.7} & \textbf{+251.1} & {55.1} & {+27.0} \\
      
      \emph{w/} SFT & 1.1k & - & 86.0  & +7.5 & 33.3 & +153.8 & 48.0 & +10.6 \\
      \midrule
       \emph{w/} SFT & 1.1k & 0.7k & 87.1 & +8.9 & 40.0 & +200.8 & 49.0 & +12.9 \\
       \emph{w/} SFT & 1.1k & 1.6k & 87.4 & +9.2 & \textbf{46.7} & \textbf{+251.1} & 53.0 & +22.1 \\
       \emph{w/} SFT (STILL-2) & 1.1k & 1.8k & 89.8 & +12.3 & 40.0 & +200.8 & \textbf{56.1} & \textbf{+29.3} \\
      \midrule
       \emph{w/} SFT \& DPO & 1.1k & 0.3k & 87.2 & +9.0 & 30.0 & +125.6 & 49.5 & +14.1 \\
       \emph{w/} SFT \& DPO (STILL-2) & 1.1k & 1.0k & 85.4 & +6.8 & \textbf{46.7} & \textbf{+251.1} & 51.0 & +17.5 \\
      \bottomrule
      \end{tabular}
      \label{tab:main_res}
\end{table}

\subsection{Main Results}

In this part, we present a detailed performance comparison of various methods on the selected evaluation benchmarks, as shown in Table~\ref{tab:main_res}. The results include performance metrics for o1-like models, general-purpose models, and several approaches based on the backbone model with additional training methods. We report both the accuracy and the gain relative to the backbone's performance.

From the table (the first part of Table~\ref{tab:main_res}), we can observe that industry-level slow-thinking reasoning systems achieve excellent performance across the three benchmarks, showing significant improvement on the most challenging benchmark, AIME. Overall, o1-preview demonstrates more balanced performance, while R1 and QwQ perform better in the math domain. These results indicate the effectiveness of slow thinking in enhancing the complex reasoning capabilities of LLMs.

Secondly, distillation-based variants of our approach (the first group in the second part of Table~\ref{tab:main_res}) can yield very competitive results, as shown in the second group of rows, approaching those of industry counterparts. For example, using 3.9k distilled  instances obtained from both R1 and QwQ after our preprocessing, our method achieves 46.7\% accuracy on AIME and 90.2\% accuracy on MATH-OAI.
We find that conducting careful data cleaning, selection, and mixing of demonstration instances is very useful when implementing this variant.  
Another observation is that increasing the amount of high-quality demonstration data can effectively improve model performance, as evidenced by the comparison between models trained with 1.1k and 3.9k instances.

Thirdly, the iteratively trained variants of our approach  (the second and third groups in the second part of Table~\ref{tab:main_res}) can also achieve promising results across the three benchmarks. 
Using the variant \emph{w/ SFT 1.1k} as a reference, we observe that incorporating exploration and self-improvement leads to performance improvements for both SFT or DPO, \eg the performance on AIME goes from 33.3\% to 40.0\%, 46.7\%, and 40.0\% respectively. Note that our variants are optimized by the iteratively refined training datasets in Section~\ref{sec-selection}. Since the exploration on challenging problems is very time-consuming, we maintain all the derived correct trajectories of multiple trained or experimented variants in a global pool across multiple runs of experiments. The explored instances (the third column of Table~\ref{tab:main_res}) are selected from this pool, rather than from a single variant or the preceding variants in the same run of experiments. Additionally, we find that using more explored instances can also lead to performance improvement to some extent.

Empirically, we  find that the improvement of iterative training is often limited to the initial iterations and might lead to performance fluctuations on some benchmarks. We speculate that, due to the constrained number of rollouts (at most 20 in our experiments), a portion of challenging problems cannot be sufficiently explored by our reasoning model—meaning the model fails to arrive at the correct answer—thereby significantly limiting its exploration capacity. As future work, we plan to extend the search time for exploration to address this limitation.

Overall, our distillation-based variant (with 3.9k instances) achieves the best performance among all our attempts, approaching the performance of industry-level reasoning systems. Meanwhile, the variants incorporating exploration and self-improvement also show substantial improvements over the backbone model.

\subsection{Further Analysis of Data Mixture}
\label{sec-IL}
During SFT training, we prepare a mixture of training data from different domains and varying difficulty levels. In this section, we examine the impact of this data mixture on the model’s performance. Specifically, our training dataset consists of three main sources: \emph{hard mathematical problems} (corresponding to difficulty levels such as AIME or the Mathematical Olympiad), \emph{normal mathematical problems} (corresponding to the MATH-OAI difficulty level), and \emph{data from other domains} (corresponding to other disciplines in GPQA). Since the math domain typically contains many challenging reasoning problems, we prioritize it as the main domain.

For the three sources, we experiment with different proportions for data mixture: \emph{w/o hard problems} (removing the hard mathematical problems), \emph{w/o other domains} (removing all non-math data), and \emph{mixed domain data} (including all three parts with a carefully tuned distribution).

We present the performance comparison in Table~\ref{tab-data_active_learning} and derive three major findings. 
First, excluding the hard problem data leads to a significant drop in performance. This highlights the importance of hard problems in enhancing the reasoning model’s capabilities, particularly on the most challenging benchmark, AIME, in our experiments. We observe that hard problems typically require a longer thought process to reach the correct solution (as indicated by the average thought length statistics), which helps better guide and teach LLMs to generate long-form thought.

Second, using mathematical data alone results in a strong performance across all three benchmarks, not limited to the math domain. This suggests that reasoning with long-form thought is an inherent capability of LLMs, which can be generalized across domains once properly elicited or taught. This finding is particularly significant for the design of generalized reasoning algorithms.

\begin{table}[t]
    \caption{Performance comparison with different mixtures for multi-domain data. We also report the average length for each data mixture. 
    }
    \small
    \centering
    \begin{tabular}{cccccc}
        \toprule
        \textbf{Settings} & \textbf{Avg. Length} & \textbf{MATH-OAI} & \textbf{AIME} & \textbf{GPQA} & \textbf{Avg.} \\
        \midrule
        \emph{w/o} hard problems & 2866  & 86.0 & 33.3 & 51.0 & 56.8\\
       \emph{w/o} other domains & 3389  & 87.4 & 46.7 & 53.0 & 62.4 \\
        {mixed} & 3162  & 89.8 & 40.0 & 56.1 & 62.0 \\
        \bottomrule
    \end{tabular}
    \label{tab-data_active_learning}
\end{table}

\begin{table}[t]
    \caption{Performance comparison of different variants with the DPO algorithm. For SFT, we optimize the model over both the parts of thought and solution. We incorporate the performance of the fine-tuned model after imitation learning as the reference. }
    \small
    \centering
    \begin{tabular}{cccccc}
        \toprule
        \textbf{DPO} & \textbf{SFT} & \textbf{MATH-OAI} & \textbf{AIME} & \textbf{GPQA} & \textbf{Avg.} \\
        \midrule
        Thought + Solution & \ding{55} & 86.2 & 33.3 & 52.5 & 57.4   \\
        Thought + Solution & \ding{52} & 87.2 & 26.7 & 43.9 & 52.6   \\
        Thought & \ding{55} & 85.4 & 46.7 & 51.0 & 61.0   \\
        Thought & \ding{52} & 87.6 & 33.3 & 50.5 & 57.1   \\
        \bottomrule
    \end{tabular}
    \label{tab-objective_function}
\end{table}

\subsection{Further Analysis of DPO Training}
\label{exp-dpo}

Another aspect to consider is the setting of the DPO algorithm in Section~\ref{sec-self_improve}. We introduce two major modifications to the original DPO algorithm: (1) aligning only the thought process, and (2) incorporating SFT for more stable optimization. To examine the impact of these strategies, we compare the performance using variants that align both the thought and the solution, as well as those that exclude the SFT loss. 

The comparison results are presented in Table~\ref{tab-objective_function}. An interesting finding is that aligning only the thought process can yield promising results. This is likely because the thought process is the core element to learn, and once it is well established, LLMs can readily generate the corresponding solution. Another observation is that the SFT loss seems to have little positive influence when aligning both the thought and solution, as the solution is already incorporated into the DPO training.

While our results appear to fluctuate across different runs, we recommend that future research conduct more thorough experiments to explore various alignment strategies when using DPO to train reasoning models.

\section{Conclusion}

In this paper, we present a detailed introduction to a reproduced o1-like reasoning system, STILL-2. We outline a three-phase development approach for implementing such a capable system, where the model is initially trained using distilled long-form thought data and then undergoes self-improvement by exploring difficult problems. Our system has demonstrated strong performance on three challenging evaluation benchmarks. We find that the slow-thinking mode can be easily transferred across domains and is particularly effective at solving hard, complex problems. Our main findings can be summarized as follows:

\begin{itemize}
\item The ability to perform long-form thinking can be effectively elicited by training with a small amount of high-quality demonstration data. Once established, this ability appears to naturally generalize across domains.

\item Demonstration data from the math domain is particularly well-suited for developing the long-form thinking ability of LLMs, and data with longer thought processes appears especially effective in enhancing the model's capacity to tackle challenging problems.

\item Unlike the formal responses generated by LLMs in a fast-thinking mode, the thought process is typically expressed in a flexible, informal manner, serving to guide LLMs toward the correct path to the solution.

\item The slow-thinking capacity can be effectively enhanced through exploration and self-improvement, whereas the improvements from offline learning methods seem to occur primarily in the initial iterations, especially for challenging tasks.

\end{itemize}

This work follows our previous study~\cite{jiang2024technical} on building complex reasoning systems. Despite the promising results, our exploration remains preliminary, and there is still a substantial capacity gap compared to industry-level systems. As future work, we plan to investigate how to scale our training approach and extend its capacity to more complex tasks. 
As always, we are committed to keeping our technical approach \emph{open}, and we welcome collaboration and support in computational resources.

\section*{Acknowledgments}
We are grateful to DeepSeek and Qwen for providing public access to their reasoning systems. We also thank Haoxiang Sun for sharing his mathematical expertise. 

\bibliographystyle{unsrt}
\bibliography{ref.bib}

\begin{table}[h]
    \centering
    \renewcommand{\arraystretch}{1.2} 
    \setlength{\tabcolsep}{8pt} 
    \begin{tabular}{p{14cm}}
        \begin{tcolorbox}[colback=gray!5, colframe=gray!40, sharp corners, width=\linewidth, boxrule=0.5mm]
        \small
        \textbf{Problem: }
        Convert the point $(0,3)$ in rectangular coordinates to polar coordinates.  \\

        Enter your answer in the form $(r, \theta),$ where $r > 0$ and $0 \leq \theta < 2 \pi$.
        \end{tcolorbox} \\[-6pt] 
        
        \begin{tcolorbox}[colback=gray!5, colframe=gray!40, sharp corners, width=\linewidth, boxrule=0.5mm]
        \small
        \textbf{Output: }
        \\
        \textit{<|begin\_of\_thought|>} \\
        \\
        Alright, so I've got this problem here: I need to convert the point (0,3) from rectangular coordinates to polar coordinates. I know that in polar coordinates, points are represented as ($r$, $\theta$), where $r$ is the radius or the distance from the origin, and $\theta$ is the angle from the positive $x$-axis. \\
        \\
        First, I need to find $r$. I remember that $r$ is the distance from the origin to the point, which can be found using the Pythagorean theorem. So, $r = \sqrt{x^2 + y^2}$. Plugging in the values, $x$ is 0 and $y$ is 3, so \\
        \[
        r = \sqrt{0^2 + 3^2} = \sqrt{9} = 3
        \]
        Okay, so $r$ is 3. \\
        \\
        Next, I need to find $\theta$. $\theta$ is the angle that the line from the origin to the point makes with the positive $x$-axis. I think I can use the arctangent function, $\theta = \arctan(y/x)$. But wait, in this case, $x$ is 0, so $y/x$ would be $3/0$, which is undefined. Hmm, that's a problem. \\
        \\
        I recall that when $x$ is 0, the point lies on the $y$-axis. If $y$ is positive, the point is on the positive $y$-axis, and if $y$ is negative, it's on the negative $y$-axis. In this case, $y$ is 3, which is positive, so the point is on the positive y-axis. \\
        \\
        I know that the angle for the positive $y$-axis is $\pi/2$ radians, or 90 degrees. So, $\theta$ should be $\pi/2$. \\
        \\
        But let me double-check. The arctangent function usually gives angles in the range of $-\pi/2$ to $\pi/2$, but since the point is on the $y$-axis, I need to consider the correct quadrant. The point (0,3) is in the upper half-plane, so $\theta$ should indeed be $\pi/2$. \\
        \\
        Also, the problem specifies that $r$ should be positive, which it is, and $\theta$ should be between 0 and $2\pi$, which $\pi/2$ is. \\
        \\
        So, putting it all together, the polar coordinates should be $(3, \pi/2)$. \\
        \\
        I think that's the answer. \\
        \\
        \textit{<|end\_of\_thought|>} \\
        \\
        \textit{<|begin\_of\_solution|>} \\
        \\
        To convert the point $(0, 3)$ from rectangular coordinates to polar coordinates, we need to find $r$ and $\theta$. \\
        \\
        1. \textbf{Calculate $r$:} \\
        \[
        r = \sqrt{x^2 + y^2} = \sqrt{0^2 + 3^2} = \sqrt{9} = 3
        \] \\
        2. \textbf{Determine $\theta$:} \\
        \\
        Since the point $(0, 3)$ lies on the positive y-axis, the angle $\theta$ is $\frac{\pi}{2}$ radians. \\
        Thus, the polar coordinates are: \\
        \[
        \boxed{(3, \frac{\pi}{2})}
        \] \\
        \textit{<|end\_of\_solution|>}
        \end{tcolorbox}
    \end{tabular}
    \caption{A sample solution to a problem in MATH-OAI.}
    \label{tab:case1}
\end{table}

\begin{table}[h]
    \centering
    \renewcommand{\arraystretch}{1.2} 
    \setlength{\tabcolsep}{8pt} 
    \begin{tabular}{p{\linewidth}}
        \begin{tcolorbox}[colback=gray!5, colframe=gray!40, sharp corners, width=\linewidth, boxrule=0.5mm]
        \small
        \textbf{Problem: }
        Every morning Aya goes for a $9$-kilometer-long walk and stops at a coffee shop afterwards. When she walks at a constant speed of $s$ kilometers per hour, the walk takes her 4 hours, including $t$ minutes spent in the coffee shop. When she walks $s+2$ kilometers per hour, the walk takes her 2 hours and 24 minutes, including $t$ minutes spent in the coffee shop. Suppose Aya walks at $s + \frac{1}{2}$ kilometers per hour. Find the number of minutes the walk takes her, including the $t$ minutes spent in the coffee shop.
        \end{tcolorbox} \\[-6pt] 
        
        \begin{tcolorbox}[colback=gray!5, colframe=gray!40, sharp corners, width=\linewidth, boxrule=0.5mm]
        \small
        \textbf{Output: }\\ \textit{<|begin\_of\_thought|>} \\
        Alright, so I've got this problem here about Aya's morning routine. She walks 9 kilometers and then stops at a coffee shop for $t$ minutes. The time she takes for the whole routine depends on her walking speed. \\
        When she walks at $s$ kilometers per hour, the total time is 4 hours, including the time at the coffee shop. When she walks at $s + 2$ kilometers per hour, the total time is 2 hours and 24 minutes, again including the coffee shop time. \\
        \\
        I need to find out how many minutes the whole routine takes when she walks at $s + \frac{1}{2}$ kilometers per hour. \\
        Okay, first things first, I need to understand what's being asked. There are two scenarios given with different speeds and total times, including the coffee shop stop. I need to find the total time for a third speed. \\
        \\
        Let's break it down. In the first scenario, walking at $s$ km/h, the total time is 4 hours, which includes the walking time and $t$ minutes at the coffee shop. In the second scenario, walking at $s + 2$ km/h, the total time is 2 hours and 24 minutes, again including $t$ minutes at the coffee shop. \\
        \\
        I need to find the total time when she walks at $s + 0.5$ km/h. \\
        First, I should probably find the value of $s$, her original walking speed. To do that, I can set up equations based on the information given. \\
        \\
        Let's denote the walking time without the coffee shop as $T$. So, in the first scenario: Walking time $T + t$ minutes = 4 hours. Similarly, in the second scenario: Walking time $T' + t$ minutes = 2 hours and 24 minutes. But I need to express $T$ and $T'$ in terms of $s$. We know that time = distance / speed. So, for the first scenario: \\
        \[
        T = \frac{9}{s} \, \text{hours}
        \]
        And for the second scenario: \\
        \[
        T' = \frac{9}{s + 2} \, \text{hours}
        \]
        So, I have two equations: \\
        \[
        \frac{9}{s} + \frac{t}{60} = 4
        \]
        \[
        \frac{9}{s + 2} + \frac{t}{60} = 2.4
        \]
        I can solve these two equations to find $s$ and $t$. \\
        Using the quadratic formula, I get: \\
        \[
        s = 2.5 \, \text{km/h}, \quad t = 24 \, \text{minutes}
        \]
        To find the total time when she walks at $s + 0.5 = 3$ km/h: \\
        \[
        \text{Total time} = \frac{9}{3} + \frac{t}{60} = 3 + 0.4 = 3.4 \, \text{hours}
        \]
        Converting 3.4 hours to minutes: \\
        \[
        3 \times 60 + 0.4 \times 60 = 204 \, \text{minutes}
        \]
        Thus, the total time is 204 minutes. \\
        \textit{<|end\_of\_thought|>} \\
        \end{tcolorbox} \\
    \end{tabular}
    \caption{A sample solution to a problem in AIME.}
    \label{tab:case2}
\end{table}

\begin{table}[h]
    \ContinuedFloat
    \centering
    \renewcommand{\arraystretch}{1.2} 
    \setlength{\tabcolsep}{8pt} 
    \begin{tabular}{p{\linewidth}}
        \begin{tcolorbox}[colback=gray!5, colframe=gray!40, sharp corners, width=\linewidth, boxrule=0.5mm]
        \small
        \textit{<|begin\_of\_solution|>} \\
        Given that Aya walks 9 kilometers at a speed of $s$ kilometers per hour, and the total time including $t$ minutes at the coffee shop is 4 hours, we have: \\
        \[
        \frac{9}{s} + \frac{t}{60} = 4
        \]
        Similarly, when she walks at $s + 2$ kilometers per hour, the total time is 2 hours and 24 minutes: \\
        \[
        \frac{9}{s + 2} + \frac{t}{60} = 2.4
        \]
        Solving for $s$ and $t$, we get: \\
        \[
        s = 2.5 \, \text{km/h}, \quad t = 24 \, \text{minutes}
        \]
        To calculate the total time at speed $s + \frac{1}{2} = 3$ km/h: \\
        \[
        \text{Total time} = \frac{9}{3} + \frac{t}{60} = 3 + 0.4 = 3.4 \, \text{hours}
        \]
        Converting to minutes: \\
        \[
        3 \times 60 + 0.4 \times 60 = 204 \, \text{minutes}
        \]
        Thus, the total time is: \\
        \[
        \boxed{204 \, \text{minutes}}
        \]
        \textit{<|end\_of\_solution|>}
        \end{tcolorbox}
    \end{tabular}
    \caption*{Table~\ref{tab:case2} (Continued): A sample solution to a problem in AIME.}
\end{table}

\end{document}